\pdfoutput=1
\documentclass[sigconf, authorversion, nonacm]{acmart}
\usepackage{times}
\usepackage{latexsym}
\usepackage[T1]{fontenc}
\usepackage[utf8]{inputenc}
\usepackage{microtype}
\usepackage{graphicx}
\usepackage{tabularx}
\usepackage{booktabs}
\usepackage{ragged2e}
\usepackage{enumitem}
\usepackage{adjustbox}
\usepackage{amsmath}

\title{Contextually Aware E-Commerce Product Question Answering using RAG}

\author{Praveen Tangarajan}
\author{Anand A. Rajasekar}
\author{Manish Rathi}
\author{Vinay Rao Dandin}
\author{Ozan Ersoy}

\affiliation{%
  \institution{Flipkart US R\&D Center}
  \city{Bellevue}
  \state{Washington}
  \country{USA}
}

\begin{document}

% \footnotetext{* Denotes primary contribution.}

\begin{abstract}
E-commerce product pages contain a mix of structured specifications, unstructured reviews, and contextual elements like personalized offers or regional variants. Although informative, this volume can lead to cognitive overload, making it difficult for users to quickly and accurately find the information they need. Existing Product Question Answering (PQA) systems often fail to utilize rich user context and diverse product information effectively. We propose a scalable, end-to-end framework for e-commerce PQA using Retrieval Augmented Generation (RAG) that deeply integrates contextual understanding. Our system leverages conversational history, user profiles, and product attributes to deliver relevant and personalized answers. It adeptly handles objective, subjective, and multi-intent queries across heterogeneous sources, while also identifying information gaps in the catalog to support ongoing content improvement. We also introduce novel metrics to measure the framework's performance which are broadly applicable for RAG system evaluations.
\end{abstract}
\maketitle

\section{Introduction}
E-commerce platforms have transformed how consumers discover and evaluate products by aggregating diverse information sources into a single product page. These pages combine structured data (e.g., specifications, pricing), unstructured content (e.g., user reviews, FAQs), and contextual elements such as personalized offers, regional variants, and localized payment options. While intended to support informed decision-making, the sheer volume and variety of this content often lead to cognitive overload particularly on mobile devices resulting in decision fatigue and a fragmented shopping experience. Conversational commerce has emerged as a promising direction to alleviate this overload by offering interactive, dialogue-based assistance. However, traditional product question answering (PQA) systems, typically based on keyword retrieval or early neural models, fall short in handling noisy data, capturing nuanced intent, or adapting to user context. These systems often ignore critical signals like prior interactions, evolving preferences, or the layered structure of product information.

The advent of Large Language Models (LLMs) and Generative AI has created new opportunities for more flexible and context-aware information access. In this work, we introduce a scalable, end-to-end framework for e-commerce PQA that leverages Retrieval Augmented Generation (RAG) combined with deep contextual modeling. Our approach integrates user interaction history, real-time query signals, and heterogeneous product data (structured, unstructured, and semi-structured) to generate accurate and personalized answers. Key innovations include its ability to maintain and leverage conversational context and mechanisms to identify information gaps in the product catalog for continuous improvement. Furthermore, we propose a robust evaluation protocol, including novel metrics focused on contextual accuracy and information completeness, which are also valuable for broader assessments of the RAG system.

The remainder of the paper is organized as follows: Section 2 reviews related work in PQA and RAG. Section 3 defines the problem. Section 4 presents our proposed framework. Section 5 reports experimental results, and Section 6 concludes with future directions.

\section{Related Work}

Product Question Answering (PQA) is a specialized QA task focused on generating responses to customer queries by leveraging diverse e-commerce data—structured specifications, unstructured reviews, and FAQs. Unlike general-purpose QA \citep{Rajpurkar2016} or domain-specific QA in biomedical \citep{Jin2023} and legal \citep{Gil2021WhatInAQuestion} domains, PQA must synthesize both factual and opinion-based content from heterogeneous sources. 

Early approaches emphasized opinion mining \citep{Moghaddam2011, Yu2012OpinionQA} and structured querying using SQL or knowledge graphs \citep{Frank2007, Tapeh2008, Li2019AliMeKG}. More recent methods leverage pre-trained language models to extract answers from unstructured content like reviews and product descriptions \citep{Gupta2019AmazonQA, Gao2019ProductAware, Zhang2020AnswerFact}, and address semi-structured data using attribute ranking and generative models \citep{Kulkarni2019, shen-etal-2022-semipqa}. The advent of Large Language Models (LLMs) and Retrieval Augmented Generation (RAG) \citep{lewis2020retrieval} has further advanced PQA. Enhancements include improved retrieval \citep{Ma2023QueryRewriting, Ilin2024SelfRetrieval,kulkarni2024reinforcement}, iterative reasoning \citep{Shao2023IterativeRAG}, and domain alignment \citep{Yang2023KnowledgeAugmentedLM}. Evaluation efforts have introduced tailored benchmarks for RAG-based systems \citep{Chen2023BenchmarkingRAG, Lyu2024RGB}.

A comprehensive survey \citep{Deng2022PQASurvey} classifies PQA methods as opinion-based, extraction-based, retrieval-based, and generation-based. While generation models offer the greatest flexibility, they remain prone to hallucination and lack robust evaluation protocols. Our work builds on these foundations by introducing a scalable, context-aware RAG framework that integrates conversational history, product grounding, and unified intent modeling. We further propose a novel evaluation protocol designed specifically for e-commerce QA scenarios.

\section{Problem Statement}
\label{sec:ps}

We define the e-commerce platform's product catalog as $\mathcal{I} = \{I_p \mid p \in \mathcal{P}\}$, where each product $p$ is associated with an information tuple $I_p = (S_p, U_p, M_p)$. These sources correspond to {structured} attributes ($S_p$), {unstructured} content like user reviews ($U_p$), and {semi-structured} entries ($M_p$).

A user interaction at turn $i$ is defined by a static or dynamic user context $U_{ctx}$ (e.g., profile, preferences) and a conversational history $H^{(i-1)} = \{(q^{(t)}, y^{(t)})\}_{t=1}^{i-1}$. Given the current query $q^{(i)}$, which may be objective, subjective, or multi-intent, our goal is to generate a response $y^{(i)}$ that is: \textbf{(1) Contextually Relevant} and personalized to $U_{ctx}$ and $H^{(i-1)}$; \textbf{(2) Factually Accurate} and grounded in the retrieved product information $\mathcal{I}_{P^{(i)}}$; and \textbf{(3) Informative about Limitations}, acknowledging information gaps when the context is insufficient.

We model this task as learning the conditional probability $P(y^{(i)} \mid q^{(i)}, H^{(i-1)}, U_{ctx}, \mathcal{I}_{P^{(i)}})$.

\section{Proposed Architecture}

\begin{figure}[hbtp]
    \centering
    \includegraphics[width=\columnwidth]{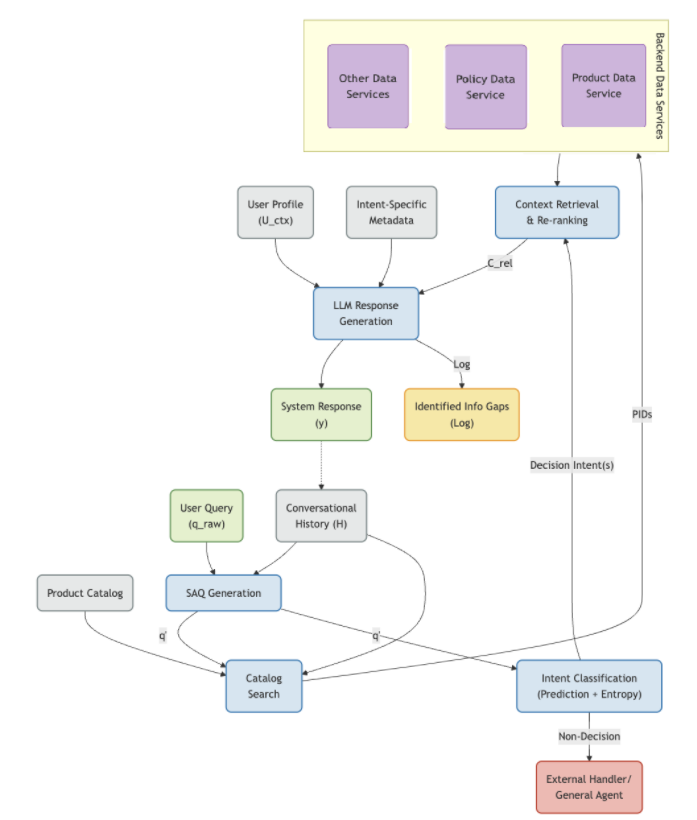}
    % Vspace with a negative value pulls the caption UP towards the image
    \vspace{-8pt} % Adjust this value to get the desired spacing
    
    \caption{Diagram illustrating the proposed query processing architecture.}
    \label{fig:sys_arch_2}
    
    % This pulls the text BELOW the figure UP towards the caption
    % Use this one carefully, as it can make things look cramped.
    \vspace{-12pt} 
\end{figure}

Our proposed context-aware framework, illustrated in Figure~\ref{fig:sys_arch_2}, processes user queries through a modular pipeline that integrates conversational history, product identification, intent understanding, and targeted retrieval.

The pipeline begins with a Standalone Query (SAQ) Module, which transforms the raw user query $q^{(i)}$ along with its conversational history $H^{(i-1)}$ into a self-contained, contextualized query $q'^{(i)}$. This is passed to a Catalog Search Model that detects product mentions and maps them to unique product IDs $P_{IDs}^{(i)}$ from the catalog $\mathcal{P}$.

Next, a unified Intent Model predicts a distribution over predefined intent classes, including a Non-Decision category for out-of-domain queries and multiple Decision intents such as specifications, offers, or payment options. If the Non-Decision intent is not dominant, the system assesses entropy over the Decision intents to guide retrieval strategy: low entropy indicates a clear dominant intent, triggering focused retrieval, while high entropy reflects ambiguity, prompting retrieval for the top-N (e.g., top three) intents to maximize coverage.

For each identified intent, the system invokes relevant backend APIs to retrieve structured attributes ($S_p$), unstructured reviews ($U_p$), semi-structured ($M_p$), and other product-related data. To manage the potentially large volume of retrieved content, a Retrieval and Re-ranking module selects and prioritizes context snippets based on semantic relevance to $q'^{(i)}$, yielding a concise set $\mathcal{C}_{rel}^{(i)}$.

Finally, the refined context $\mathcal{C}_{rel}^{(i)}$ and query $q'^{(i)}$ are composed into a tailored prompt for a Large Language Model (LLM). The LLM synthesizes this information to generate a coherent, fluent, and contextually faithful response $y^{(i)}$, with built-in mechanisms to acknowledge any information gaps.

% % Use figure* instead of figure to span both columns
% % Use figure* to span both columns
% \begin{figure*}[htbp]
%     \centering
%     % Replace L, B, R, T with estimated trim amounts (e.g., 10pt 20pt 10pt 5pt)
%     \includegraphics[width=\textwidth, trim=0pt 0pt 0pt 0pt, clip]{emnlp2023-latex/sys_arch_2.png}
%     \caption{Diagram illustrating the proposed query processing architecture.}
%     \label{fig:sys_arch_2}
% \end{figure*}

% \begin{figure}[htbp]
%     \centering
%     \includegraphics[width=\columnwidth]{emnlp2023-latex/sys_arch_2.png}
%     \caption{Diagram illustrating the proposed query processing architecture.}
%     \label{fig:sys_arch_2}
% \end{figure}

% \begin{figure}[hbtp]
%     \centering
%     \includegraphics[width=\columnwidth]{emnlp2023-latex/sys_arch_2.png}
    
%     % Vspace with a negative value pulls the caption UP towards the image
%     \vspace{-8pt} % Adjust this value to get the desired spacing
    
%     \caption{Diagram illustrating the proposed query processing architecture.}
%     \label{fig:sys_arch_2}
    
%     % This pulls the text BELOW the figure UP towards the caption
%     % Use this one carefully, as it can make things look cramped.
%     \vspace{-12pt} 
% \end{figure}

\begin{table}[hbtp]
\centering
\small
\setlength{\tabcolsep}{4pt}
\renewcommand{\arraystretch}{1.1}
\begin{tabularx}{\columnwidth}{@{}>{\RaggedRight\arraybackslash}X >{\RaggedRight\arraybackslash}X >{\RaggedRight\arraybackslash}X@{}}
\toprule
\textbf{Scenario} & \textbf{Conversation Snippet} & \textbf{SAQ Output} \\
\midrule

\textbf{Follow-up on same product} &
\textit{On iPhone 13 product page} \newline
\textbf{U:} Battery size? \newline
\textbf{B:} \textit{<answer>} \newline
\textbf{U:} Display size? &
What is the display size of iPhone 13? \\
\addlinespace

\textbf{Switch to a new product} &
\textit{On iPhone 13 product page} \newline
\textbf{U:} Battery size? \newline
\textbf{B:} \textit{<answer>} \newline
\textbf{U:} How about iPhone 14? &
What is the battery size of iPhone 14? \\
\addlinespace

\textbf{Product → Accessory search} &
\textit{On iPhone 13 product page} \newline
\textbf{U:} Battery size? \newline
\textbf{B:} \textit{<answer>} \newline
\textbf{U:} Show me cases for this phone. &
Show me cases for iPhone 13. \\
\addlinespace

\textbf{Browsing → Specific product} &
\textit{On Browse page} \newline
\textbf{U:} Show 2 door refrigerators. \newline
\textbf{B:} \textit{<shows multiple fridges including LG 242 L Frost Free 2 Star>} \newline
\textbf{U:} Capacity of LG fridge? &
What is the capacity of LG 242 L Frost Free 2 Star? \\
\bottomrule
\end{tabularx}
\caption{Illustrative SAQ Scenarios and Outputs (User and Bot turns are abbreviated for space)}
\label{tab:saq_scenarios}
% \vspace{2pt}
% {\footnotesize \textbf{U:} User \quad \textbf{B:} Bot}
\end{table}

\subsection{Standalone Query (SAQ)}
The Standalone Query (SAQ) module is the crucial first component in our pipeline, responsible for enabling multi-turn conversation. Its primary function is to rewrite the current user query ($q^{(i)}$) into a self-contained, contextualized query ($q'^{(i)}$) by incorporating the conversational history ($H^{(i-1)}$). This process serves two key purposes: (1) resolving co-references (e.g., pronouns like "it") and (2) performing product disambiguation by mapping ambiguous references (e.g., "the second one") to their canonical product names based on the preceding context. The resulting query, $q'^{(i)}$, encapsulates all necessary information, allowing downstream components to operate without needing access to the full conversation history. Table~\ref{tab:saq_scenarios} illustrates how the SAQ module handles diverse conversational scenarios.

\subsection{Catalog Search}
The Catalog Search module maps product mentions within the contextualized query ($q'^{(i)}$) to unique platform Product IDs. It employs a hierarchical strategy, prioritizing an \textbf{(1) exact match} of extracted product names against the conversational history ($H^{(i-1)}$). If this fails, it proceeds to \textbf{(2) fuzzy matching} against both the history and the broader platform catalog to handle variations or typos. As a final fallback, \textbf{(3) the most salient name} from $q'^{(i)}$ is used to query external and platform search APIs to retrieve a canonical product name, which is then mapped to its ID. This cascaded approach ensures robust disambiguation by prioritizing in-session context before broadening the search.

\subsection{Intent Model}
User query intent is determined by a \textbf{unified BERT-based multi-class classification model}, which categorizes the contextualized query ($q'^{(i)}$) into predefined classes. The two primary categories are: \textbf{\textit{non\_decision}} for out-of-scope queries (e.g., general search like queries) that are routed outside the PQA pipeline, and fine-grained \textbf{\textit{decision}} intents for queries seeking specific product information (e.g., specifications, offers).

The model outputs a probability distribution over these intents, and an entropy-based mechanism handles ambiguity. First, if \textit{non\_decision} is predicted with high confidence, the query is routed out. Otherwise, the entropy of the distribution over \textit{decision} intents is analyzed: low entropy selects the single, dominant intent, while high entropy selects the top-N most probable intents to handle multi-faceted queries. The selected \textit{decision} intent(s) then dictate which APIs are invoked to fetch relevant data. This unified approach simplifies the system architecture while enabling efficient and nuanced handling of diverse user queries.

\subsection{Retrieval}
Upon determining the decision intent(s), the framework initiates a two-stage retrieval process. In \textbf{Stage 1: API Orchestration}, the system calls backend APIs corresponding to the selected intent(s), fetching broad data from canonical sources spanning structured attributes, semi-structured FAQs, and unstructured content. To handle multi-intent queries, data for all top-predicted intents is retrieved to ensure coverage.

Since these API outputs are often verbose and contain irrelevant information, \textbf{Stage 2: Context Reduction and Re-ranking} is performed. This stage employs a bi-encoder Semantic Textual Similarity (STS) model \citep{reimers-2019-sentence-bert} to score and rank chunks of the retrieved data against the contextualized query $q'^{(i)}$. The top-N chunks with the highest semantic similarity are selected to form a final, concise context, $\mathcal{C}_{rel}^{(i)}$. This reduced context is critical for improving the generative model's factual accuracy and mitigating hallucinations.

\subsection{Generation}
Once the relevant context $\mathcal{C}_{rel}^{(i)}$ is prepared, a strategically ordered prompt is constructed to guide the LLM. The prompt begins with \textbf{(1) System Persona and Core Instructions}, which set the model's behavior through static guidelines (e.g., strict grounding on the context, no speculation, informal tone) and dynamic, intent-specific few-shot examples that uses advanced prompting like Chain-of-Thought. Next, \textbf{(2) The Retrieved Reduced Context} ($\mathcal{C}_{rel}^{(i)}$) provides the factual basis, anchored by the product title (from $P_{IDs}^{(i)}$) to prevent information leakage, followed by the top-ranked data snippets. To enable personalization, \textbf{(3) The User Context} ($U_{ctx}$) is then included, deliberately placed after the factual data to prioritize grounding. For complex queries, dynamic \textbf{(4) Intent-Specific Metadata} is added to explain platform-specific terms. Finally, \textbf{(5) The Contextualized User Query} ($q'^{(i)}$), already refined by the SAQ model, is appended. This structured composition enables the LLM to produce responses that are grounded, accurate, personalized, and engaging, balancing data fidelity with a seamless user experience.

\section{Experimentation and Results}

\subsection{Approach}
Our experimental methodology followed a rigorous, iterative process integrating prompt engineering, automated LLM-based evaluation, and human alignment.

Each system module was first defined with a clear task specification, followed by constructing an initial generation prompt incorporating a system persona and curated few-shot examples using GPT-3.5 \cite{openai2023gpt35}. We used proprietary e-commerce data spanning categories like electronics and fashion, tailored to each module’s requirements. Parallelly, we defined precise evaluation metrics and developed a structured evaluation prompt using GPT-4 \cite{openai2023gpt4} to automatically assess outputs over a representative dataset sampled from production. A subset of outputs was manually annotated by human labelers using the same definitions to validate and iteratively refine the evaluation prompt until its judgments closely matched human feedback.
Once aligned, we iteratively improved the generation prompts based on automated evaluation scores. Final validation used a fresh dataset, with both automated and human evaluations ensuring reliability. After deployment, production data was continuously collected and leveraged to fine-tune lightweight LLMs (e.g., 2B parameter models) and train downstream classifiers critical to system operations. This end-to-end approach blending prompt design, automated LLM feedback, human verification, and model tuning enabled the development of a robust, efficient, and scalable framework.

\subsection{Results}

\subsubsection{SAQ}

The Standalone Query (SAQ) module transforms user utterances into fully contextualized queries for downstream processing. Built on a fine-tuned in-house LLaMA 3–8B model, SAQ resolves co-references and disambiguates product mentions, enabling components to operate independently of full conversational history.

To evaluate SAQ, we use three metrics: (1) overall query restructuring accuracy, (2) turn-1 accuracy, and (3) turn except-1 accuracy (multi-turn without first query). Outputs are assessed using GPT-4–based evaluation prompts aligned with human judgments, showing under 1\% deviation.

We began with GPT-3.5 using prompt engineering, but after reaching its performance ceiling, we transitioned to fine-tuning in-house models, ultimately scaling up to LLaMA 3–8B, our current in-house production model. This final model surpasses 95\% turn except-1 accuracy with sub-500ms latency, essential for production use.

As shown in Table~\ref{tab:saq-metrics}, our production model achieves 97.60\% overall accuracy, 99.08\% accuracy on first-turn queries, and 95.93\% on multi-turn scenarios. These results underscore SAQ’s ability to reliably generate context-aware queries essential to the system’s performance.

\begin{table}[hbtp]
    \centering
    \scriptsize
    \begin{adjustbox}{max width=\linewidth, center}
        \begin{tabular}{|l|c|c|c|}
            \hline
            \textbf{Model} & \textbf{Overall (\%)} & \textbf{Turn 1 (\%)} & \textbf{Turn >1 (\%)} \\
            \hline
            GPT-3.5 (0125)              & 94.31 & 98.68 & 89.41 \\
            In-house Phi-2              & 95.36 & 98.45 & 91.90 \\
            In-house Phi-3              & 95.57 & 98.33 & 92.48 \\
            In-house LLaMA 3–8B         & \textbf{97.60} & \textbf{99.08} & \textbf{95.93} \\
            \hline
        \end{tabular}
    \end{adjustbox}
    \caption{Performance of SAQ Model Variants}
    \label{tab:saq-metrics}
\end{table}

\subsubsection{Catalog Search}
The Catalog Search Model demonstrates decent performance in identifying the correct full product name from SAQ output, as well as from provided products with pre-stored PIDs in the conversational context. For this task, we employ a GPT-3.5-based prompt iteratively refined through continuous evaluations with human labeling. This refined process yields a high accuracy rate of 90.03\%. However, we observed that product name hallucinations occur in 0.75\% of cases, while existing product names are missed 6.39\% of the time.

In instances where the extracted product name is absent from the context, fuzzy matching is employed against the platform's catalog to select the best matching product and retrieve its corresponding Product ID. This multi-step approach effectively enhances the model's accuracy and reliability, solidifying its integral role in the pipeline.

% \subsubsection{Intent Model}
% The Intent Model, a crucial component for routing and tailoring downstream processing, employs a BERT-based multi-class classification architecture. To assess its efficacy, we utilized standard classification metrics: Precision, Recall, and F1-score, calculated on a per-intent basis. An iterative development process was adopted, focusing on identifying and improving underperforming classes based on these metrics to enhance the model's overall discriminative power.
% The performance metrics presented in Table~\ref{tab:intent-metrics} reflect the model's ability to predict the single, most probable intent (top-1). It is important to note that in our deployed system, as described in Section 4, we utilize an entropy-based mechanism to select the top-k intents when faced with ambiguity. This top-k approach demonstrably improves practical system performance and user satisfaction by accommodating multi-faceted queries or less confident predictions, although the results below focus on the foundational top-1 classification accuracy.
% The model achieves a high overall top-1 accuracy of 93.17\% and a robust weighted average F1-score of 92.57\%. This indicates strong performance across the majority of query volume, which is critical for the system's effectiveness in a production e-commerce environment.
\subsubsection{Intent Model}
Our Intent Model, a BERT-based multi-class classifier, routes queries and tailors downstream processing. It achieves 93.17\% top-1 accuracy and a 92.57\% weighted F1-score (Table~\ref{tab:intent-metrics}), a performance level critical for our production environment. While these metrics reflect top-1 classification, our deployed system uses an entropy-based mechanism to select the top-k intents for ambiguous queries, which demonstrably improves practical performance and user satisfaction.

\begin{table}[hbtp] % Added table environment, placement options
\centering % Added centering

\scriptsize % Matched target table
\begin{adjustbox}{max width=\linewidth, center} % Matched target table adjustbox
    % Using target format with | and \hline
    \begin{tabular}{|l|c|c|c|}
        \hline
        \textbf{Intent Class} & \textbf{Precision (\%)} & \textbf{Recall (\%)} & \textbf{F1-score (\%)} \\ % Added units to headers
        \hline
        non\_decision         & 81.72\%            & 61.29\%         & 70.05\%           \\ % Kept %
        authenticity          & 80.00\%            & 92.31\%         & 85.71\%           \\ % Kept %
        checkout              & 99.26\%            & 99.41\%         & 99.33\%           \\ % Kept %
        delivery\_sla          & 93.83\%            & 99.66\%         & 96.66\%           \\ % Kept %, used \_
        offers\_and\_discounts & 94.78\%            & 97.99\%         & 96.36\%           \\ % Kept %, used \_
        payment\_options       & 96.65\%            & 99.43\%         & 98.02\%           \\ % Kept %, used \_
        product\_exchange      & 97.01\%            & 97.01\%         & 97.01\%           \\ % Kept %, used \_
        product\_spec          & 83.92\%            & 95.59\%         & 89.38\%           \\ % Kept %, used \_
        return\_policy         & 98.93\%            & 100.00\%        & 99.46\%           \\ % Kept %, used \_
        size\_and\_fit         & 92.86\%            & 47.27\%         & 62.65\%           \\ % Kept %, used \_
        stock\_availability    & 96.69\%            & 96.69\%         & 96.69\%           \\ % Kept %, used \_
        variant               & 56.25\%            & 54.55\%         & 55.38\%           \\ % Kept %
        warranty              & 93.58\%            & 97.77\%         & 95.63\%           \\ % Kept %
        \hline % Added hline before averages like target style implies separation
        \textbf{Macro Avg}    & 89.65\%            & 87.61\%         & 87.87\%           \\ % Kept %
        \textbf{Weighted Avg} & 93.77\%            & 94.78\%         & 94.01\%           \\ % Kept %
        \hline
    \end{tabular}
\end{adjustbox}
\caption{Intent Model Performance Across Classes} % Adjusted caption slightly
\label{tab:intent-metrics} % Added label
\end{table}

\subsubsection{Context Reduction}

To provide concise and relevant inputs to the LLM generator, we explored several context reduction strategies, evaluating them using \textbf{Recall@k}—the proportion of queries where the ground truth answer appeared among the top-$k$ retrieved segments.

Supervised Deep Passage Retrieval (DPR) models achieved strong recall but were impractical for dynamic e-commerce catalogs due to the need for frequent retraining. Unsupervised alternatives like fastText \citep{bojanowski2017enriching} offered better scalability but with significantly lower recall.

We ultimately adopted a bi-encoder Semantic Textual Similarity (STS) model inspired by Sentence-BERT. It selects the top-$k$ most relevant sentences from the retrieved context. Crucially, we domain-adapted this model using a large in-house e-commerce FAQ dataset and trained it with a triplet loss objective:
\begin{equation}
\resizebox{1.0\linewidth}{!}{
    $\displaystyle L = \sum_{i} \max(0, ||f(q_i) - f(p_i)||^2 - ||f(q_i) - f(n_i)||^2 + \alpha)$
}
\label{eq:triplet_loss}
\end{equation}
where $f(\cdot)$ is the bi-encoder embedding function and $(q_i, p_i, n_i)$ denote query, positive, and negative samples, with $\alpha$ as the margin.
As shown in Table~\ref{tab:cr-perf}, this STS model achieved a Recall@k of 98.32\%, outperforming fastText (95.06\%) and slightly exceeding DPR (98.10\%). For example, setting $k=15$ for the fashion category retrieved ground truth in 98\% of cases. This approach balances high recall with scalability, adapting seamlessly across product categories and updates without retraining.
\begin{table}[hbtp] % Added placement options
\centering
\scriptsize % Matched target table
\begin{adjustbox}{max width=\linewidth, center} % Matched target table adjustbox
    % Using target format with | and \hline
    \begin{tabular}{|l|c|}
    \hline
    \textbf{Context Reduction Model} & \textbf{Recall @ Top-k (\%)} \\ % Added units
    \hline
    fastText \citep{bojanowski2017enriching} & 95.06\% \\ % Added %
    Supervised DPR & 98.10\% \\ % Added %
    \textbf{Bi-encoder STS (Ours)} & 98.32\% \\ % Added %
    \hline
    \end{tabular}
\end{adjustbox}
\caption{Performance comparison of context reduction algorithms.} % Adjusted caption slightly
\label{tab:cr-perf} % Moved label after caption
\end{table}
\subsubsection{Generation}
The generation module in our RAG framework produces accurate, complete, and contextually grounded responses to factual, subjective, and multi-intent queries, strictly within the retrieved context $\mathcal{C}_{rel}^{(i)}$. Informational gaps are handled responsibly: if context is insufficient, the system abstains from speculation and returns ``IDK'' (I Don’t Know), avoiding hallucinated or misleading answers. Missing context does not imply feature absence; if inference is needed, responses are cautious and cite evidence where possible.

To ensure faithfulness and context-awareness, we define a robust evaluation framework assessing answer quality (factuality, completeness, precision) and alignment with context. Because generation quality depends on context adequacy, we also evaluate the retrieved context.

\noindent\textit{Assumptions:} For each query $q'^{(i)}$, human or LLM-based judgment determines if $\mathcal{C}_{rel}^{(i)}$ sufficiently supports the query or its sub-intents. An answer $y^{(i)}$ is complete if it addresses all answerable components or returns ``IDK'' for unanswerable parts.

\noindent\textbf{Note:} ``IDK'' represents pattern of responses generated when the retrieved context does not contain sufficient information to answer a query.

\noindent\textbf{I. Context Quality Metric} \\
\textbf{Context Coverage (CCov):} Assesses whether the retrieved context provides sufficient information to fully or partially answer the query. \textit{Metric:} Percentage of queries at least partially answerable from the retrieved context.

\noindent\textbf{II. Answer Evaluation Metrics} \\
We evaluate generated answers based on four distinct scenarios: (S1) context is sufficient and an answer is provided; (S2) context is insufficient and the system responds with ``IDK''; (S3) context is insufficient but the system attempts a factual answer (not IDK); and (S4) context is sufficient but the answer is ``IDK''. Key metrics include \textit{Factuality/Faithfulness}, which measures whether answers are factually correct and grounded in the retrieved context, computed as $\text{Grounded Accuracy} = \frac{N_{S1,FC}}{N_{S1}}$. \textit{Answer Completeness} assesses if all sub-questions are answered, given by $\text{Completeness} = \frac{N_{S1,Comp}}{N_{S1}}$.
\textit{Precision} quantifies the fraction of valid factual answers among all attempted (S1 and S3): $\text{Precision} = \frac{N_{S1,Good}}{N_{S1} + N_{S3}}$. \textit{Recall} measures the fraction of correctly answered, truly answerable queries (S1 and S4): $\text{Recall} = \frac{N_{S1,Good}}{N_{S1} + N_{S4}}$. \textit{Accuracy} reflects overall correctness, rewarding both valid S1 answers and correct S2 IDK responses: $\text{Accuracy} = \frac{N_{S1,Good} + N_{S2}}{M}$, where $M$ is the total number of queries."Good" requires both factual correctness AND completeness. Finally, the \textit{Hallucination Rate} captures the proportion of unsupported or incorrect answers (S3 and incorrect S1), calculated as $\text{Hallucination Rate} = \frac{N_{S3} + N_{S1,Bad}}{M}$.

\noindent Table~\ref{tab:answer-gen-aggregated} summarizes generation performance across these metrics and user intent types, highlighting the system’s strengths in managing contextual ambiguity and mitigating hallucinations.

\begin{table}[hbtp] % Added placement options
\centering
\scriptsize % Kept from original
\begin{adjustbox}{max width=\linewidth, center} % Added center option for consistency
    % Using target format with | and \hline
    \begin{tabular}{|l|c|c|c|c|}
    \hline
    \textbf{Intent} & \textbf{Coverage (\%)} & \textbf{Precision (\%)} & \textbf{Recall (\%)} & \textbf{Hallucination (\%)} \\ % Added units
    \hline
    authenticity          & 94.0\% & 98.2\% & 98.4\% & 1.69\% \\ 
    checkout              & 85.9\% & 98.0\% & 99.5\% & 1.32\% \\ 
    delivery\_sla          & 92.4\% & 96.6\% & 98.4\% & 3.11\% \\ % Used \_
    offers\_and\_discounts & 90.85\% & 96.42\% & 98.62\% & 2.90\% \\ % Used \_
    payment\_options       & 95.9\% & 96.5\% & 97.6\% & 3.36\% \\ % Used \_
    product\_exchange      & 79.4\% & 91.3\% & 92.9\% & 6.89\% \\  % Used \_
    product\_spec          & 86.4\% & 97.2\% & 98.7\% & 1.93\% \\  % Used \_
    return\_policy         & 84.0\% & 96.7\% & 96.9\% & 2.72\% \\  % Used \_
    size\_and\_fit         & 35.2\% & 90.3\% & 94.8\% & 3.42\% \\  % Used \_
    stock\_availability    & 51.4\% & 95.1\% & 97.1\% & 2.50\% \\  % Used \_
    variant               & 77.6\% & 94.4\% & 95.8\% & 4.38\% \\
    warranty              & 90.0\% & 97.7\% & 98.7\% & 2.00\% \\ \hline
    \end{tabular}
\end{adjustbox}
\caption{Answer Generation Performance Across Intents: Coverage, Precision, Recall, and Hallucination Rate} % Adjusted caption slightly
\label{tab:answer-gen-aggregated}
\end{table}

\section{Conclusion}
In this paper we presented a scalable, end-to-end framework for e-commerce PQA that overcomes information overload by deeply integrating user and product context into a Retrieval-Augmented Generation (RAG) architecture. Our system integrates conversational history, user intent, and heterogeneous data sources to deliver personalized answers to objective, subjective, and multi-intent queries. Its core contributions include a context-aware query rewriter (SAQ model), an entropy-based intent model for disambiguation, and a two-stage retrieval process with a domain-adapted bi-encoder that ensures factual grounding and minimizes hallucinations. To validate our approach, we introduced a novel suite of metrics for evaluating RAG systems, which confirmed the architecture's effectiveness in handling information gaps with principled, transparent responses.
Deployed in a production conversational assistant, our framework serves over 5 million monthly active users, delivering an 8\% increase in user thumbs-up rates and measurable improvements in conversion and customer satisfaction.

\section{Limitations}
Our current framework operates through a modular but fixed sequential pipeline. While this design ensures efficiency and reliability for well-defined queries, it lacks flexibility in dynamically re-planning or self-correcting when faced with ambiguity, unexpected inputs, or partial failures.
A second limitation lies in the predefined intent taxonomy, which constrains the system’s ability to generalize to novel or complex user queries particularly those involving compositional reasoning or intents not captured during training. As a result, the system performs best on predictable, high-frequency queries and less effectively on the long tail of diverse user needs.
In future we plan to address these limitations through the development of a more adaptive, agentic architecture capable of dynamically selecting and sequencing tools based on real-time query analysis. Additionally, we aim to build automated pipelines for catalog enrichment by detecting and resolving information gaps, thereby improving the completeness and quality of product data over time.

\bibliography{anthology,custom}
\bibliographystyle{acl_natbib}
\appendix
\end{document}